\pdfoutput=1
\documentclass[11pt]{article}

\usepackage{emnlp2021}

\usepackage{times}
\usepackage{latexsym}

\usepackage[T1]{fontenc}
\usepackage[utf8]{inputenc}

\usepackage{microtype}

\usepackage{graphicx}
\usepackage{amsmath}
\usepackage{amssymb}
\usepackage{booktabs}
\usepackage{array}
\usepackage{enumitem}

\newcommand{\indep}{\rotatebox[origin=c]{90}{$\models$}}

\usepackage{xcolor}

\title{Text as Causal Mediators: Research Design for Causal Estimates of Differential Treatment of Social Groups via Language Aspects}

\author{Katherine A.~Keith, Douglas Rice, and Brendan O'Connor \\
University of Massachusetts Amherst \\
\texttt{kkeith@cs.umass.edu,drrice@legal.umass.edu,brenocon@cs.umass.edu}
}

\begin{document}
\maketitle

\begin{abstract}
Using observed language to understand interpersonal interactions is important in high-stakes decision making. We propose a causal research design for observational (non-experimental) data 
to estimate the natural direct and indirect effects of social group signals (e.g.~race or gender) on speakers' responses with separate aspects of language as causal mediators. We illustrate the promises and challenges of this framework via a theoretical case study of the effect of an advocate's gender on interruptions from justices during U.S.~Supreme Court oral arguments. 
We also discuss challenges conceptualizing and operationalizing causal variables such as gender and language that comprise of many components, and we articulate technical open challenges such as temporal dependence between language mediators in conversational settings. 
\end{abstract}

\section{Introduction}
Interactions between individuals are key components of social structure \cite{hinde1976social}. 
While we rarely have access to individuals' internal thoughts during these interactions, we often can observe the language they use. 
Using observed language to better understand interpersonal interactions is important in high-stakes decision making---for instance, judges' decisions within the United States legal system \cite{dnm2012power} or police interaction with citizens during traffic stops \cite{voigt2017language}.
In these settings, analysts may be interested in understanding the behavior of decision makers as individuals or at the subgroup or aggregate level.

\begin{figure}[t]
\centering
\includegraphics[width=0.8\columnwidth]{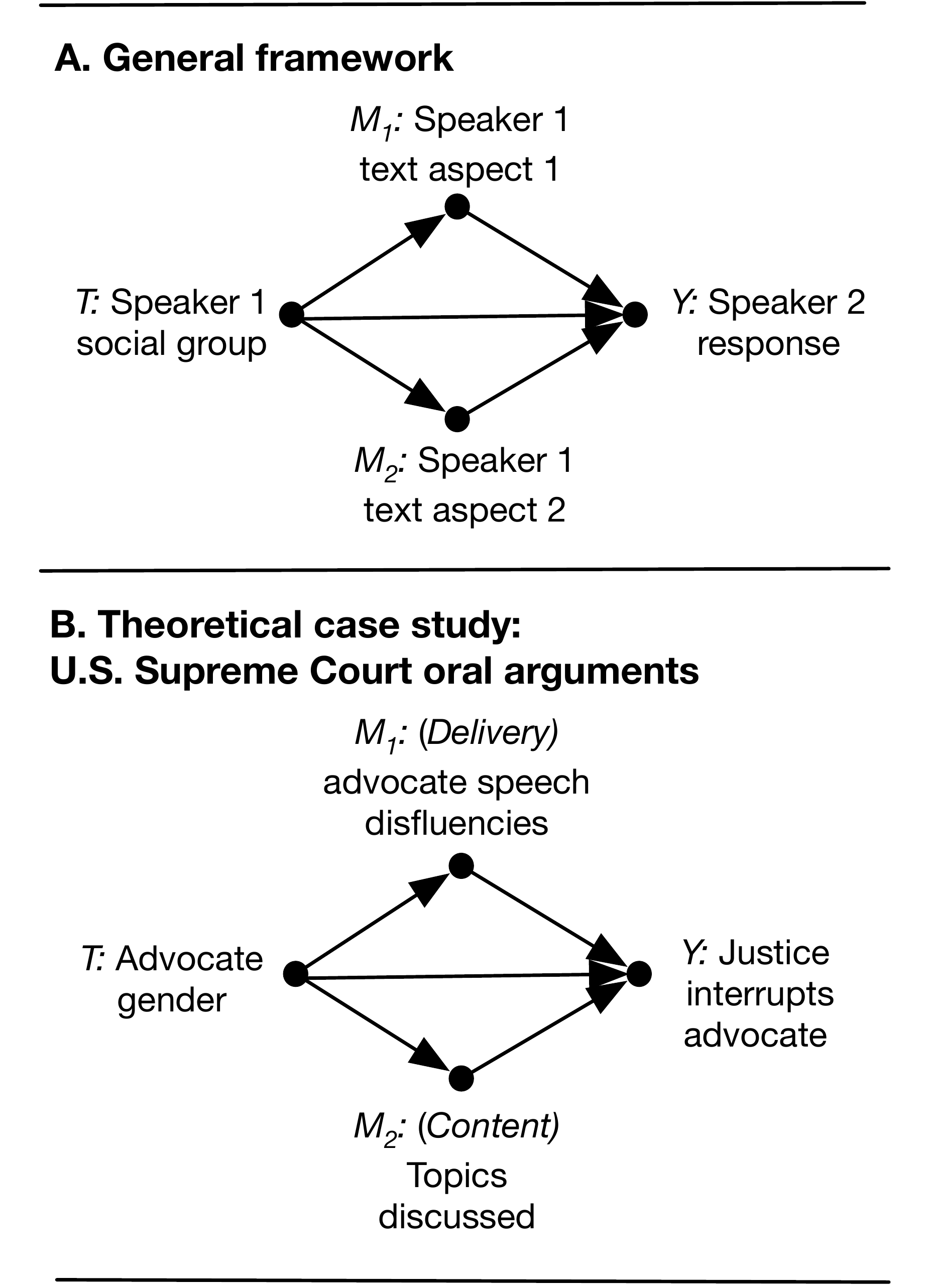}
\caption{Causal diagrams in which nodes are random variables and arrows denote causal dependence for \textbf{A.} proposed general framework for \emph{differential treatment of social groups via language aspects} and 
\textbf{B.} instantiation of the framework for a theoretical case study of U.S.~Supreme Court oral arguments.
In both diagrams, $T$ is the treatment variable, $Y$ is the outcome variable, and $M$ are mediator variables. This is a simplified schema; see Fig.~\ref{f:constitutive-diagram} for an expanded diagram.
\label{f:text-mediator-fig}}
\end{figure}

Important decision makers sometimes treat some social groups (e.g.~women, racial minorities, or ideological communities) differently than others \cite{gleason2020beyond}. Yet, quantitative analyses of this problem often do not account for all possible mechanisms that could induce this differential treatment. For instance, one might ask,
\emph{During U.S.~Supreme Court oral arguments, is a justice interrupting female advocates more because of their gender, because of the content of the advocates' legal arguments, or because of the advocates' language delivery} (Fig.~\ref{f:text-mediator-fig}B)? Accounting for these language mechanisms could help separate and estimate the remaining ``gender bias'' of justices. 
% \emph{In a conversation between Speakers 1 and 2, is Speaker 2 responding differently to Speaker 1 because Speaker 2 perceives Speaker 1 to be part of a particular social group? Or because of \emph{what} Speaker 1 says (content) or \emph{how} Speaker 1 says it (delivery)?}

We reformulate the previous question as a general \emph{counterfactual} query \cite{pearl2009causality,morgan2015counterfactuals} about two speakers: \emph{How would Speaker 2 respond if the signal they received of Speaker 1's social group flipped from A to B but Speaker 1 still used language typical of social group A?} Here, our question is about the direct causal effect of \emph{treatment}---Speaker 1's signaled social group---on \emph{outcome}---Speaker 2's response---that is not through the causal pathway of the \emph{mediator}---an aspect of language (Fig.~\ref{f:text-mediator-fig}A).\footnote{See \S\ref{subsec:treatment} for a discussion on when and how social groups (e.g.~gender or race) can be used as causal treatments.}

The fundamental problem with this and any counterfactual question is that we cannot go back in time and observe an individual counterfactual 
% (Speaker 2's perception of Speaker 1's social group flipping from A to B) 
while holding all other conditions the same \cite{holland1986statistics}. Furthermore, in many high-stakes, real-world settings (e.g.~the U.S.~Supreme Court), we cannot run experiments to randomly assign treatment and approximate these counterfactuals. Instead, in these settings, causal estimation must rely on \emph{observational} (non-experimental) data.

In this work, we focus on this observational setting and build from causal mediation methods \cite{pearl2001direct,imai2010general,vanderweele2016mediation} to specify a research design of causal estimates of \emph{differential treatment of social groups via language aspects}.
Other work has used causal mediation analysis to better understand components of natural language processing (NLP) models \cite{vig2020investigating,finlayson2021causal}.
However, this work is more closely aligned with studies that focus on causal estimation in which text is one or more causal variables \cite[e.g.,][]{veitch2020adapting,roberts2020adjusting,keith2020text,zhang2020quantifying,pryzant-etal-2021-causal}.

Our focus is on the research design, and we therefore intentionally do not present empirical results. Instead, we discuss the potential promises and challenges of this causal research design with both general examples and concrete examples from a theoretical case study of U.S.~Supreme Court arguments. This aligns with \citet{rubin2008objective} who argues ``design trumps analysis'' in observational studies and emphasizes the importance of conceptualizing a study before any outcome data is analyzed.  
% We address critiques of our design in \S\ref{subsec:treatment} and \S\ref{sec:challenges} including: potential threats to validity when using social groups as a causal treatment, temporal dependence between language mediators in conversational settings, dependence between multiple language mediators, and dependence between perception of social groups and linguistic perception.

Overall, we make the following contributions:  
% \vspace{-0.2cm}
% \begin{itemize}
\begin{itemize}[leftmargin=0.3cm]
    \item We propose a new causal research design to estimate the natural indirect and direct effects of social group signal on speakers' responses with separate aspects of language as causal mediators (\S\ref{sec:identification}).  
    \item We illustrate the promises and challenges of this framework via a theoretical case study of the effect of an advocate's gender on interruptions by justices during U.S.~Supreme Court oral arguments. (\S\ref{sec:case-study}).
    \item We discuss challenges researchers might face conceptualizing and operationalizing the causal variables in this research design (\S\ref{sec:operationalization}).
    \item We directly address critiques of using social groups (e.g.~race or gender) as treatment and construct gender and language as \emph{constitutive} variables, building from \citet{sen2016race,hu2020s} (\S\ref{subsec:treatment} and \S\ref{subsec:lang-mediators}).
    % \item We also define gender and language as \emph{constitutive} variables (\S\ref{subsec:treatment};\S\ref{subsec:lang-mediators}) in response to recent calls for more engagement in the ontological assumptions of using social groups (e.g.}~race or gender) as treatment in causal inference \cite{sen2016race,kohler2018eddie,hanna2020towards,hu2020s,kasirzadeh2021use}.
    \item We articulate potential open challenges in this research design including temporal dependence between mediators in conversations, causal dependence between multiple language mediators, and dependence between social group perception and language perception (\S\ref{sec:challenges}). 
\end{itemize}

% \kkcomment{Switch ``bias'' to ``differential treatment.''}

% \kkcomment{``framework'' or ``research design'' everywhere?} 

% \emph{Is there gender bias? Is there racial discrimination?} This paper takes a \emph{causal inference} approach and addresses these questions via \emph{counterfactuals}. 

% Specifically, this paper articulates a causal research design for observational (non-experimental) data that untangles multiple aspects of language as causal mediators in order to address counterfactual questions. For example,  
% \emph{In a dyadic conversation between Speaker 1 and Speaker 2, is Speaker 2 responding to Speaker 1 based on \emph{what} Speaker 1 says (content), \emph{how} Speaker 1 says it (delivery), or non-language pathways?}

% We walk the reader through the conceptual points necessary to set-up an empirical study of this form. We believe a rigorous theoretical and conceptual framework is especially important in light of recent calls for more engagement on the ontological assumptions of using social groups (e.g.~race or gender) as treatment in causal inference or studies of algorithmic fairness \cite{sen2016race,kohler2018eddie,hanna2020towards,hu2020s,kasirzadeh2021use}

\begin{table*}[t]
  \centering
%   \resizebox{0.98\linewidth}{!}{ %makes it fit within the margin limits
{\small
      \begin{tabular}{ >{\raggedright\arraybackslash}p{15.5cm}}
    %   \toprule
    %   Ex. & \\
      \toprule
      \textbf{\emph{(A) Case: Kennedy v. Plan Administrator for DuPont Sav. and Investment Plan (2008-07-636)}} \\
      \textcolor{gray}{\textbf{Mark Irving Levy:} [...] The QDRO provision is an objective checklist that is easy for -- for plan administrators to follow.} \\
      \textcolor{gray}{\textbf{Antonin Scalia:} What if they had agreed to the waiver apart from [...] We'd be in the same suit that you're - - that you say we have to avoid, wouldn't we?} \\
      \textbf{Mark Irving Levy:} \textcolor{blue}{I don't think so.} \textcolor{blue}{I mean I think} that would be an alienation. \\
      \textbf{Antonin Scalia:} Well, if it's an alienation, but his point is that a waiver is not an alienation. \\ %2008_07-636
      \hline
      \textbf{\emph{(B) Case: Lozano v. Montoya Alvarez (2013-12-820)}} 
      \\
      \textcolor{gray}{\textbf{Ann O'Connell Adams:} Well - -} \\
      \textcolor{gray}{\textbf{Antonin Scalia:} I mean, it seems to me it just makes that article impossible to apply consistently country to country.} \\
      \textbf{Ann O'Connell Adams:} - - No, \textcolor{blue}{I don't think so.} \textcolor{red}{And - - and}, the other signatories \textcolor{red}{have - - have} almost all,  \textcolor{blue}{I mean I think} the Hong Kong court does say that it doesn't have discretion, but it said in that case nevertheless it would, even if it had discretion, it wouldn't order the children returned. But the other courts of signatory countries that have interpreted Article 12 have all found a discretion, whether it be in Article 12 or in Article 8. And if I - - \\
      \textbf{Antonin Scalia:} Have they exercised it? Have they exercised it, that discretion which they say is there?  \\ %"2013_12-820"
\bottomrule 
  \end{tabular}
  }
  \caption{
  Selected utterances from the oral arguments of two U.S.~Supreme Court cases, A \cite{oyez2021kennedy} and B \cite{oyez2021lozano}, with advocates Mark Irving Levy (male) and Ann O'Connell Adams (female) respectively. Justice Antonin Scalia responds to both advocates.  
  Hedging language is highlighted in \textcolor{blue}{blue}. 
  Speech disfluencies are highlighted in \textcolor{red}{red}. \textcolor{gray}{Gray-colored} utterances directly proceed the target utterances (non-gray colored) in the oral arguments. \label{t:paired-example}}
\end{table*}

% \kkcomment{emphasize that we are particularly interested in interpretable indirect effects through multiple aspects of language}! 

% \textbf{Our contributions:}

\section{Theoretical Case Study: Gender Bias in U.S.~Supreme Court Interruptions}\label{sec:case-study}

To motivate our causal research design and illustrate challenges that arise with it, we focus on a specific theoretical case study---the effect of advocate gender on justice interruptions via advocates' language during United States Supreme Court oral arguments (Fig.~\ref{f:text-mediator-fig}B).
% The United States Supreme Court is the apex of the American legal and judicial professions. 
% Thus, there is enormous interest in quantifying bias in individual justices' decision making and the effects of group identity on court outcomes \kkcomment{cites?}. 
% If measurable gender, racial, or ideological bias is found in this elite, highly-educated segment of the population, it is provocative for examining these forces throughout society.
The substantive motivation for this theoretical case study is built from previous work examining the role of interruption and gender on the Court. \citet{patton2017lawyer} found female lawyers are interrupted earlier in oral arguments, allowed to speak for less time, and subjected to longer speeches by justices; \citet{jacobi2017justice} found female justices are interrupted at disproportionate rates by their male colleagues; and \citet{gleason2020beyond} found justices are more likely to vote for the female advocate's side when the female advocate uses emotional language.
% \citet{jacobi2017justice} found female justices are interrupted at disproportionate rates by their male colleagues, and 
% \citet{gleason2020beyond} found justices are more likely to vote for a female advocate's party when the female advocate uses emotional language.

\textbf{Counterfactual questions.} We present a novel causal approach to understanding gender bias in Supreme Court oral arguments that corresponds to the following counterfactual questions:   
% \begin{enumerate}[leftmargin=0.3cm,noitemsep]
\begin{enumerate}
    \item \emph{(NDE)}: How would a justice's interruptions of an advocate change if the signal of the advocate's gender the justice received flipped from male to female, but the advocate still used language typical of a male advocate?   
    \item \emph{(NIE)}: How would a justice's interruptions of an advocate change if a male advocate used language typical of a female advocate but the signal of the advocate's gender the justice received remained male? 
\end{enumerate}
which we show correspond to the \emph{natural direct effect} (NDE) and \emph{natural indirect effect} (NIE) respectively in \S\ref{sec:identification}. In \S\ref{sec:operationalization}, we walk through the theoretical conceptualization and empirical operationalization of advocate gender (treatment), interruption (outcome), and advocate language (mediators). 

\textbf{Intuitive example.} 
We describe intuitive challenges of our causal research design by contrasting Examples A and B in Table~\ref{t:paired-example}.
Levy---a male advocate---is not interrupted by Justice Antonin Scalia, but Adams---a female advocate---is interrupted \citep{oyez2021kennedy,oyez2021lozano}.
\emph{Why was the female advocate interrupted?} \emph{Was it because of her gender or because of \emph{what} she said or \emph{how} she said it}?
We hypothesize one causal pathway between gender and interruption is through the mediating variable hedging---expressions of deference or politeness.\footnote{Previous work has shown hedging is used more often by women \cite{lakoff1973language,poos2002cross}, and we hypothesize judges might respond more positively to more authoritative language (less hedging) from advocates.} Suppose we operationalize hedging as certain key phrases, e.g.~``I don't think so'' and ``I mean I think.'' An initial causal design might assign a binary hedging indicator to utterances and then compare average interruption outcomes for male and female advocates conditional on the hedging indicator.

However, advocate utterances matched on this hedging indicator could have a number of latent mediators and confounders. In Table~\ref{t:paired-example}, Adams has speech disfluencies (``and - - and'' and ``have - - have'' shown in red) which might cause Scalia to get frustrated and interrupt. The cases are from different areas of the law,\footnote{The Supreme Court Database codes Ex.~A as ``economic activity'' and Ex.~B as ``civil rights'' \cite{spaeth2021supreme}.} and Scalia may interrupt more during cases that are in areas he has more personal interest. The advocate utterance in Ex.~B is longer (more tokens) and longer utterances may be more likely to be interrupted. In Ex.~B, Scalia interrupts Adams just prior to the target utterance which possibly indicates a more ``heated'' portion of the oral arguments during which interruptions occur more on average. With these confounding and additional mediator challenges, a simple causal matching approach (e.g.~\citet{stuart2010matching,roberts2020adjusting}) is unlikely to work and we advocate for the causal estimation strategy presented in \S\ref{subsec:estimation}.
We move from this case study to a formalization of our causal research design in \S\ref{sec:identification}.
% for the causal adjustment with modeling (\S\ref{subsec:estimation}). 

% Give examples of a speech disfluency. 

% Give examples of gendered topic. 

% \textbf{Data.} 

\section{Causal Mediation Formalization, Identification, and Estimation}\label{sec:identification}
Many causal questions involve \emph{mediators}---variables on a causal path between treatment and outcome. For example, what is the effect of gender\footnote{See \S\ref{sec:operationalization} for discussion of operationalizing difficult causal variables such as gender.} (treatment) on salary (outcome) with and without considering merit (a mediator)? If one intervenes on treatment, 
% $do(\text{gender})$, 
then one would activate both the ``direct path'' from gender to salary \emph{and} the ``indirect path'' from gender through merit to salary. Thus, a major focus of causal mediation is specifying conditions under which one can separate estimates of the \emph{direct effect} from the \emph{indirect effect}---the former being the effect of treatment on outcome \emph{not} through mediators and the later the effect through mediators. 

We use this causal mediation approach to formally define our framework. For each unit of analysis (see \S\ref{subsec:unit}), $i$, let $T_i$ represent the treatment variable---the social group, e.g.~gender of an advocate---and $Y_i$ represent the outcome variable---the second speaker's response, e.g.~a judge's interruption or non-interruption of an advocate. For each defined mediator $j$, let $M^j_i$ represent the mediating variable---an aspect of language, e.g.~an advocate's speech disfluencies or the topics of an utterance. Let $X_i$ represent any other confounders between any combination of the other variables. 
% ($T$, $M$, and $Y$).

%Since causal mediation ,
%the \emph{do}-notation formalism is insufficient ;
%\footnote{In the words of \citet{pearl2001direct}, a mediation research question ``cannot be represented in the standard syntax of $do(x)$ operators---it does not involve fixing any of the variables in the model but, rather, modifying the causal paths in the model.''} 

We use the potential outcomes framework \cite{rubin1974estimating} to define the natual direct and indirect effects.\footnote{\citet{pearl2001direct} notes \emph{do}-notation cannot represent causal mediation questions, since they concern counterfactual paths, not interventions of variables.} Let $M_i(t)$ represent the (counterfactual) potential value the mediator would take if $T_i=t$. Then $Y_i(t, M_i(t'))$ is a doubly-nested counterfactual that represents the potential outcome that results from both $T_i=t$ and potential value of the mediator variable with $T_i=t'$. With this formal notation, we define the individual \emph{natural direct effect (NDE)} and \emph{natural indirect effect (NIE)}:\footnote{\citet{pearl2016causal} defines the NDE and NIE in terms of the non-treatment condition, $T=0$. Others (e.g.~\citet{imai2010general} and \citet{van2011targeted}) give alternate definitions of these quantities in terms of $T=1$. We follow \citeauthor{pearl2016causal}'s definitions in the remainder of this work.} 
    \begin{align} \label{eqn:nde-def}
        \text{NDE}_i&=Y_i(1, M_i(0)) - Y_i(0, M_i(0)) \\ \label{eqn:nie-def}
        \text{NIE}_i&= Y_i(0, M_i(1)) - Y_i(0, M_i(0))
    \end{align}
These correspond to the two counterfactual questions from \S\ref{sec:case-study} if $T_i=0$ and $T_i=1$ represent the gender signal of the advocate being male and female respectively.

\subsection{Estimands}
We second the advice of \citet{lundberg2021your} and recommend researchers explicitly state their estimand of interest.  
As we briefly touch on in the introduction, some studies may be interested in the estimand as the \emph{individual}-level natural direct and indirect effects (Equations~\ref{eqn:nde-def} and \ref{eqn:nie-def}). For example, a legal scholar may be interested in an individual U.S.~Supreme Court case and estimate the individual NIE and NDE for this single case in order to evaluate how ``fair'' the case was with respect to the gender of an advocate. Machine learning approaches to estimating individual-level causal effects are promising \cite{shalit2017estimating} but may not be applicable to all datasets. In contrast, more feasible---and potentially equally substantively valid---estimands may be at the \emph{subgroup} level (e.g.~effects of all cases about civil rights or all cases for a particular justice) or aggregate level. Here, the estimands are some kind of aggregation over Equations~\ref{eqn:nde-def} and \ref{eqn:nie-def}. Thus, in Section~\ref{subsec:estimation}, we provide estimators for general population-level (\emph{not} individual-level) estimands.  
    
\subsection{Interpretation of the NDE as ``bias''} \label{subsec:interpret-nde}
Many applications of causal mediation aim to quantify ``implicit bias'' or ``discrimination'' via the natural direct effect. 
However, if all relevant mediators are not accounted for, one cannot interpret the estimand of the natural direct effect as the actual direct causal effect \citep[p.135]{van2011targeted}.
Nevertheless, if we separate the total effect into the proportion that is the NDE and the NIE with the mediators to which we have access, our analysis moves \emph{closer} to estimating the true direct effect between treatment and outcome. Thus, in this work we emphasize the value of having interpretable mediators (i.e.~language aspects) for which the NIE is a meaningful quantity to analyze in itself. 

\subsection{Identification}
Like any causal inference problem, we first examine the \emph{identification assumptions} necessary to claim an estimate as causal. The key assumption particular to causal mediation is that of \emph{sequential ignorability} \cite{imai2010general}:
\begin{enumerate}[leftmargin=0.4cm]
    \item Potential outcomes and mediators are independent of treatment given confounders 
    \begin{equation}
        \{Y_i(t',m), M_i(t) \} \indep T_i \mid X_i=x
    \end{equation}
    \item Potential outcomes are independent of mediators given treatment and confounders 
    \begin{equation} 
    Y_i(t', m) \indep M_i(t) \mid \{T_i=t, X_i=x\}
    \end{equation}
\end{enumerate}
for $t, t' \in \{0, 1\}$ and all values of $x$ and $m$. 
% \bto{`Outcomes' or `Potential outcomes'?}

% \noindent
\emph{Mediator Independence Assumption:}\footnote{This is similar to the assumptions \citet{pryzant-etal-2021-causal} make for linguistic properties of text as treatment.}
For our particular framework, we make an additional assumption that for each language aspect we study, the mediators are independent conditional on the treatment and confounders 
\begin{equation}\label{eqn:mediator-indep}
    \forall j, j':\ 
    M^j_i(t) \indep M^{j'}_i(t) \mid \{T_i=t, X_i=x\}
\end{equation}
With this assumption, we can estimate the NIE and NDE of each mediator successively, ignoring the existence of other mediators. \cite{imai2010general,tingley2014mediation}. We discuss the validity of this assumption in \S\ref{sec:challenges}. 

These assumptions correspond to the causal relationships of a graph similar to Fig.~\ref{f:text-mediator-fig}, with the addition of confounder $X$ as a parent of all $T$, $M^j$, and $Y$ (to be more precise may require a richer formalism; e.g.\ \citet{richardson2013single}).
% Given \emph{sequential ignorability} is satisfied (along with other standard causal identification assumptions such as overlap, SUTVA etc.; see \citet{morgan2015counterfactuals}), one can use the following theorem for non-parametric identification of any counterfactual mediated outcomes:  \cite{imai2010general}: 
% \begin{align} 
% \begin{split} \label{eqn:all-pot-outcomes}
% &Y_i((t), M_i(t')) | X_i = x\\
% &=\int_{\mathcal{M}} f(Y_i | M_i=m, T_i = t, X_i=x) \\
% &dF_{M_i}(m | T_i=t', X_i=x) 
% \end{split}
% \end{align} 

\subsection{Estimation}\label{subsec:estimation}
Given the satisfaction of sequential ignorability, mediator independence, and other standard causal identification assumptions,\footnote{Overlap, SUTVA etc.; see \citet{morgan2015counterfactuals}.} we propose using the following estimators of population-level natural direct and indirect effects for each mediator $j$ \cite{imai2010general,pearl2016causal}:
\par\nobreak
{\small
\begin{align}
\begin{split} \label{eqn:nde-estimate}
&\text{SA-NDE}^j = 
\\ & \frac{1}{N} \sum_{i=1}^N \sum_{x \in \mathcal{X}} \sum_{m \in \mathcal{M}^j}
\bigg(\hat{f}^j(Y | M_i^j=m, T_i=1, X_i=x) \\
&- \hat{f}^j(Y|M_i^j=m, T_i=0, X_i=x) \bigg)
\hat{g}^j(m | T_i=0, X_i = x) \\
%\hat{g}^j(m | T_i=0, X_i = x) \\ %% BTO: these 3 lines experimental, don't use
%&\bigg(\hat{f}^j(Y | m, 1, x)
%- \hat{f}^j(Y|m,0,x) \bigg)
\end{split}
\end{align}
\begin{align}
\begin{split} \label{eqn:nie-estimate}
&\text{SA-NIE}^j = \\
&
\frac{1}{N} \sum_{i=1}^N \sum_{x \in \mathcal{X}} \sum_{m \in \mathcal{M}^j}
\hat{f}^j(Y | M^j_i =m, T_i=0, X_i = x) \\
&\bigg( \hat{g}^j(m | T_i=1, X_i =x) - \hat{g}^j(m |T_i=0, X_i =x)
\bigg)
\end{split}
\end{align}
}%
\noindent
Each is a \textbf{S}ample \textbf{A}verage estimate from $N$ data points, relying on models
trained to predict mediator and outcome given confounders and treatment:
$\hat{g}^j$ infers
mediator $j$'s probability distribution,
while $\hat{f}^j$ infers the expected outcome conditional on mediator $j$.
The estimators marginalize over confounders and mediators from their respective domains ($x \in \mathcal{X}$, $m \in \mathcal{M}^j$), which for our discrete variables is feasible with explicit sums (see \citeauthor{imai2010general} for the continuous case). %use Monte Carlo simulation to marginalize continuous mediators.
% Eqns.~\ref{eqn:nde-estimate} and \ref{eqn:nie-estimate} do with exact sums since they are discrete.

\textbf{Model fitting.}
When fitting models $\hat{f}$ and $\hat{g}$, we recommend using a cross-sample or cross-validation approach in which one part of the sample  is used for training/estimation ($S_{\text{train}}$) and the other is used for testing/inference ($S_{\text{test}}$) in order to avoid overfitting \cite{chernozhukov2017double,egami2018make}. 
With text, one must also fit a model for the mediators conditional on text, $h(m| \text{text})$ using $S_{\text{train}}$. In some cases, such as measuring advocate speech disfluencies, $h$ may be a simple deterministic function. However, when using NLP and other probabilistic models (e.g~topic models or embeddings), $h$ could be a difficult function to fit and have a certain amount of measurement error. A major open question is whether to jointly fit $h$ and $g$ at training time as advocated by previous work \cite{veitch2020adapting,roberts2020adjusting} or if  $h$ and $g$ should be treated as separate modules. At inference time, we do not use the inference text from $S_{\text{test}}$ since Eqns.~\ref{eqn:nde-estimate} and \ref{eqn:nie-estimate} only rely on the mediators through estimates from $\hat{g}$. 

\section{Conceptualization and Operationalization of Causal Variables}\label{sec:operationalization} 

For any causal research design---and particularly those in the social sciences---there are often challenges \emph{conceptualizing} the theoretical causal variables of interest. 
Even after these theoretical concepts are made concrete, 
there are often multiple ways to \emph{operationalize} these concepts.
% and it is important to assess \emph{measurement validity}---how well operationalizations match theoretical concepts \cite{adcock2001measurement}. 
We discuss conceptual and operational issues for our both our general research design and our theoretical case study. In particular, we recommend researchers formalize variables such as gender and language as \emph{constitutive} variables made of multiple components (Fig.~\ref{f:constitutive-diagram}) as per \citet{hu2020s}, or \citet{sen2016race}'s ``bundle of sticks.''

\subsection{Unit of analysis}\label{subsec:unit}
As with most causal research designs, one starts by conceptualizing the \emph{unit of analysis}---the smallest unit about which one wants to make counterfactual inquiries.
In our framework, the \emph{unit of analysis} is a certain amount of language
% \footnote{Our examples use text, but one could also use audio signals.} 
($L$) between speakers of two categories: the first category of speakers, $P_1$, are those belonging to a group of interest (e.g.~advocates) for which treatment values (e.g.~female and male) will be assigned; and the second, $P_2$, is the set of decision-makers responding to the first speakers (e.g.~judges).
% ,  
%  \begin{equation}
%  U_i=(L_i^{j\in{S_1}}, L_i^{k \in {S_2}})
%  \end{equation}
%  where $i$ indexes the unit, $j$ indicates the specific individual who is a part of $S_1$, and $k$ indicates the specific individual who is a part of $S_2$. 
 
\textbf{Operationalizations.} There are several possible operationalizations of $L$: pairs of single utterances---whenever a person from $P_1$ speaks and a person from $P_2$ responds; a thread of several utterances between persons from $P_1$ and $P_2$ within a conversation; or the entire conversation between persons from $P_1$ and $P_2$. In \S\ref{sec:challenges}, we note that selecting the unit of language could have implications for modeling temporal dependence between mediators. 

\subsection{Treatment}\label{subsec:treatment}  
At the most basic level, \emph{treatment}, $T$, in our research design is \emph{the social group} of persons in $P_1$ (Fig.~\ref{f:text-mediator-fig}). However, inspired by the \emph{causal consistency} arguments from \citet{hernan2016does},\footnote{
\emph{Consistency} is the condition that for observed outcome $Y$ and treatment $T$, the potential outcome equals the observed outcome, $Y(t)=Y$ for each individual with $T=t$. \citet{hernan2016does} presents eight versions of treatment for the causal question ``Does water kill?" to illustrate the deceptiveness of this apparently simple consistency condition. \citeauthor{hernan2016does} points out that ``declaring a version of treatment sufficiently well-defined is a matter of agreement among experts based on the available substantive knowledge'' and is inherently (and frustratingly) subjective.
} we examine several competing versions of treatment for our theoretical case study of U.S.~Supreme Court oral arguments and explain the reasons we eventually choose version~\#5 (in bold):
\begin{enumerate}[leftmargin=0.5cm]
\itemsep0em
    \item Do judges interrupt at different rates based on an advocate's \emph{gender}? 
    \item Based on an advocate's \emph{biological sex assigned at birth}? 
    \item An advocate's \emph{perceived gender}?  
    \item An advocate's  \emph{gender signal}? 
    \textbf{\item An advocate's  \emph{gender signal} as defined by (hypothetical) manipulations of the advocate's clothes, hair, name, and voice pitch?} 
    \item An advocate's \emph{gender signal} by (hypothetical) manipulations of their entire physical appearance, facial features, name, and voice pitch?
    \item An advocate's \emph{gender signal} by setting their physical appearance, facial features, name, and voice pitch to specific values (e.g.~all facial features set to that of the same 40-year-old, white female and clothes set to a black blazer and pants).  
\end{enumerate}

In critique of treatment version \#1, most social groups (e.g.~gender or race) reflect highly contextual social constructs \cite{sen2016race,kohler2018eddie,hanna2020towards}. For gender in particular, researchers have shown social, institutional, and cultural forces shape gender and gender perceptions \cite{deaux1985sex,west1987doing}, and thus viewing gender as a binary ``treatment'' in which individuals can be randomly assigned is methodologically flawed. In critique of version \#2, \emph{biological sex assigned at birth} is a characteristic that is not manipulable by researchers and the ``at birth'' timing of treatment assignment means all other variables about the individual are post-treatment. Thus, researchers have warned against estimating the causal effects of these kinds of ``immutable characteristics'' \cite{berk2005statistical,holland2008causation}. 

\citet{greiner2011causal} propose overcoming the issues in versions \#1 and \#2 by shifting the unit of analysis to the \emph{perceived gender} of the decision-maker (\#3) and defining treatment assignment as the moment the decision-maker first perceives the social group of the other individual. \citet{hu2020s} critique this \emph{perceived gender} variable and emphasize that we, as researchers, cannot actually change the internal, psychological state of decision-makers, but rather we can change the \emph{signal} about race or gender those decision-makers receive (\#4). 
However, as \citet{sen2016race} discuss, defining treatment as the \emph{gender signal} (\#4) is dismissive of the many components that make up a social construct like gender. Instead, \citeauthor{sen2016race} recommend articulating the specific variables one would potentially manipulate. For \emph{gender} in our case study, this could mean hypothetical manipulations of an advocate's dress, name, and voice pitch (\#5). 

Shifting from versions \#5 to \#6 and \#7, we define treatment in terms of more specific manipulations. However, we also enter the realm of \citeauthor{hernan2016does}'s argument that precisely defining the treatment never ends, and some aspects of \#6 and \#7 are impossible to manipulate in real-world settings such as the U.S.~Supreme Court. What does it mean to manipulate an advocate's ``entire physical appearance?''\footnote{Would justices have to interact with advocates through a computer-mediated system in which one could customize avatars of the advocates? We note, using computer-mediated avatars to signal social group identity has been used effectively in other causal studies, e.g.~\citet{munger2017tweetment}.} 
When we define treatment very specifically---e.g.~using the same 40-year old white woman as the treatment for ``female advocate'' (\#7)---are we estimating a causal effect of gender \emph{in general}?
Thus, we back-off from versions \#6 and \#7, and advocate using \#5 as our definition of treatment. 

\begin{figure*}[t]
\centering
\includegraphics[width=0.7\textwidth]{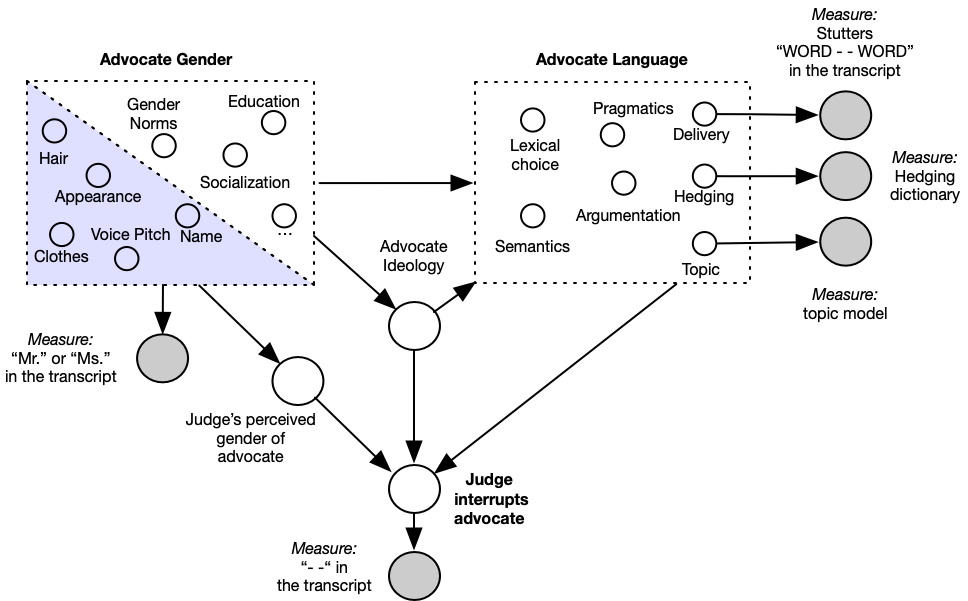}
\caption{\emph{Constitutive} causal diagram for gendered interruption in U.S.~Supreme Court oral arguments. Latent theoretical concepts are unshaded circles and observed operationalizations (measurements) of concepts are shaded circles. We provide alternative operationalizations in the text. The causal variables \emph{gender} and \emph{language} are represented as dashed lines around their constituent parts, building from the arguments of \citet{sen2016race,hu2020s}. The shaded portion of \emph{gender} consists of the gender variables that one could potentially manipulate in a hypothetical intervention.  \label{f:constitutive-diagram}}
\end{figure*}

\textbf{Constitutive causal diagrams.} With these considerations, drawing a causal diagram in which a \emph{gender} is represented as a single node seems flawed. Instead, building from \citet{sen2016race} and \citet{hu2020s}, we represent treatment (the social group) as cloud of components (a \emph{constitutive} variable), some of which are latent, some observable, and some manipulable. In Fig.~\ref{f:constitutive-diagram}, we shade the ``outward'' components of \emph{gender}---hair, appearance, clothes, voice pitch, and name---that are our hypothetical manipulations and would influence the latent variable of a judge's perceived gender of the advocate. Other ``background'' components of gender---gender norms, education, and socialization---are the components that could causally influence language.

\textbf{Case study operationalizations.} 
Even after selecting version \#5 as our conceptualization of treatment, there are still multiple operationalizations for our theoretical case study:  

\textbf{Treatment operationalization 1:} Previous work operationalizes gender in Supreme Court oral arguments by using norm that the Chief Justice introduces an advocate as ``Ms.'' and ``Mr." before their first speaking turn \cite{patton2017lawyer,gleason2020beyond}. The advantage of this operationalization is that it is simple, clean, and consistent, and occurs directly before an advocate's first utterance.\footnote{The treatment assignment timing is potentially important for the rest of the causal diagram. If we can define \emph{gender signal} and thus latent \emph{perceived gender} as happening right before an advocate first speaks, and it is not adapted or updated by the judge over the course of the oral arguments, then we can eliminate the causal arrow between variables ``language" and ``perceived gender.''}

\textbf{Treatment operationalization 2:} Alternatively, one could focus on even more specific components of gender for (hypothetical) manipulations. For instance, \citet{chen2016perceived} and \citet{chen2019attorney} measure voice pitch when studying gender on the U.S.~Supreme Court. While being more cumbersome to measure, this operationalizes gender as a real-valued (instead of binary) variable and thus potentially measures more subtle gender biases.  

\subsection{Outcome}
In our general framework, we define the \emph{outcome}, $Y$, as \emph{the response of the second speaker} (Fig.~\ref{f:text-mediator-fig}A), and we intentionally leave this variable vague and domain-specific. However, if making the leap from \emph{differential treatment} to claiming \emph{discrimination} or \emph{bias}, conceptualizing a causal outcome requires normative commitments and a moral theory of what is harmful \cite{kohler2018eddie,blodgett2020language}. 
In our case study, we conceptualize the outcome variable as a judge interrupting an advocate. This outcome is of substantive interest because, in general, interruptions can indicate and reinforce status in conversation \cite{mendelberg2014gender}, and, specifically to the U.S.~Supreme Court, justice's behavior in oral arguments has been connected to case outcomes. 

\textbf{Outcome operationalization 1:} Previous work uses the transcription norm of a double-dash (``- -'') at the end of a advocate utterance when a justice interrupts in the next utterance \cite{patton2017lawyer}. However, the validity of this operationalization relies on consistent transcription standards.

\textbf{Outcome operationalization 2:} An alternative operationalization could classify interruptions into positive (agreeing with the first speaker's comment), negative (disagreeing, raising an objection, or completely changing the topic), or neutral categories \cite{stromer2007measuring,mendelberg2014gender}. While estimating the effects of only negative interruptions could further refine the causal question---\emph{Do justices negatively interrupt female advocates more?}---this operationalization could also introduce measurement error since it could prove difficult difficult to design an accurate NLP classifier for this task. 

\subsection{Language Mediators}\label{subsec:lang-mediators} 

Our framework explicit focuses on \emph{language as a mediator} in differential treatment of social groups. 
% Reiterating \S\ref{subsec:interpret-nde}, it is often difficult to interpret the NDE as ``bias'' or ``discrimination;" however, we have estimate interpretable, substantively meaningful NIE with effective conceptualization and operationalization of language variables. 
Yet, language consists of multiple levels of linguistic structure \cite{bender2013linguistic,bender2019linguistic}, so as with social groups (\S\ref{subsec:treatment}), it is a variable that is non-modular and we believe it should be represented as constituent parts (Fig.~\ref{f:constitutive-diagram}).  

\textbf{Mediator Operationalizations:} We focus on three potential language aspects for our Supreme Court case study: (A) \emph{hedging}---expressions of deference or politeness---with an operationalization as lexical matches from a single-word hedging dictionary (e.g.~\citet{prokofieva2014hedging}); (B) \emph{speech disfluencies}---repetitions of syllables, words, or phrases---which we operationalize as the transcript noting a repeated unigram with a double dash, ``\emph{word} - - \emph{word}''; and (C) semantic \emph{topics} operationalized as a topic model \cite{blei2003latent} applied to utterances. 

\textbf{Recommendations.} 
We discuss the choice of these particular language aspects, $M^j$, for our case study as well as general recommendations for researchers operationalizing language as a mediator.

\begin{itemize}[leftmargin=0.3cm] 
\item 
Is $M^j$ \emph{interpretable}? Is there a \emph{hypothetical manipulation}\footnote{To be precise, the \emph{controlled direct effect} is the estimand in which the mediator is manipulated, $do(M)$ \cite{pearl2001direct}. In contrast, the \emph{natural} direct and indirect effects are counterfactuals on paths. However, we still find thinking through potential manipulations is helpful in refining the conceptualization of a language aspect.} of $M^j$? In contrast to prior work that treats language as a black-box in causal mediation estimates \cite{veitch2020adapting}, we advocate for using interpretable aspects of language. If language mediators are interpretable, then the NIE is both meaningful (see \S\ref{subsec:interpret-nde}) and potentially more fine-grained (we can estimate an NIE for each aspect of language that we are studying instead of a black-box approach that lumps all text into one effect). Furthermore, since identification is essential to claiming an estimate is causal and identification can only be verified qualitatively and through domain expertise, interpretable text mediators will be much easier to evaluate. 

\item 
Is there \emph{substantive theory} for causal pathways $T \rightarrow M^j$ and from $M^j \rightarrow Y$? Without such theory, studying certain aspects of language is not meaningful. For example, see \S\ref{sec:case-study} for our theoretical reasoning about the causal dependence between gender, hedging, and interruption.

\item 
To what extent does one expect \emph{measurement error} of $M^j$ when using automatic NLP tools? Our operationalizations of hedging lexicons and speech disfluencies are deterministic; however, topic model inferences are probabilistic and sensitive to changes in hyperparameters and pre-processing decisions \cite{schofield2017pulling,denny2018text}. These kinds of measurement errors are still open questions although there is recent work that examines  measurement error when text is treatment \cite{wood2018challenges}.

\item 
Is $M^j$ \emph{causally independent} from other measured language aspects, $M^{j'}$? If not, our proposed estimator from \S\ref{subsec:estimation} is invalid. Thus, one must scrutinise which aspects of language are separable and thus able to be included in the causal analysis---e.g. we could include content (topics) versus delivery (speech disfluencies) since one could hypothetically modify one without affecting the other. We discuss this assumption further in \S\ref{sec:challenges}.

\end{itemize}

% \begin{table*}[t]
%   \centering
% %   \resizebox{0.98\linewidth}{!}{ %makes it fit within the margin limits
%       \begin{tabular}{l >{\raggedright\arraybackslash}p{14cm}}
%     %   \toprule
%     %   Ex. & \\
%       \toprule
%       C &\textbf{Troy N. Giatras:}  Well, it -- it -- the ATF regulation denotes that you should even put it down if it's going to be an assault and-- \\
%       &\textbf{Antonin Scalia:} I don't understand what you've said. \\ %2008_07-608 \\
%       \hline
%       D &\textbf{Christopher M. Curran:}  Well -- well, I don't know.
% But under -- and I don't know if this is directly responsive to your question, [...]
% So I submit --\\
% &\textbf{Ruth Bader Ginsburg:} I thought you were really -- I thought it was a given that you were contesting the retroactivity only of punitive damages, not the basic cause of action, not the compensatory damages, not pain and suffering, not solatium, I -- and now you seem to be waffling or reneging on that concession. \\
% &\textbf{Christopher M. Curran:} No.
% So -- so -- so here -- here's my analysis, right? [...] \\ %case_id = "2019_17-1268"
% \bottomrule 
%   \end{tabular}
% %   }
%   \caption{Speech disfluency examples \kkcomment{may cut!} \label{t:X}}
% \end{table*}

\subsection{Non-language Mediators} 
Returning to \S\ref{subsec:interpret-nde}, there is often a tendency to interpret the NDE as something like ``pure'' \emph{gender bias}---What is the effect of gender on interruption when all other possible causal pathways are stripped away? Conceptualizing and operationalizing language aspects as mediators (\S\ref{subsec:lang-mediators}) moves the NDE towards the desired ``gender bias.'' However, there may be other mediator pathways that explain these effects. For example, in our case-study, two additional mediators of interest are advocate ideology (e.g.~liberal or conservative) and the level of ``eliteness'' of the advocate's law firm. A major validity issue is the \emph{causal independence} of these mediators from the language mediators. For instance, ideology could influence certain aspects of language (topic), and ``eliteness'' of the advocate's law firm could be a proxy for level of training which could influence the advocate's delivery.

\section{Challenges and Threats to Validity}\label{sec:challenges} 

We discuss additional challenges and threats to validity for our research design that should be addressed before implementing the design and claiming the estimates from the design are causal. 

\textbf{Temporal dependence of utterances.}
So far, we have assumed the ``units of analysis'' of text are independent (\S\ref{subsec:unit}). However, previous utterances in a conversation often influence the target utterances. For our case study, if Judge A interrupted Advocate B often in $t'<t$, interruption at $t$ is more likely (the two speakers are possibly in a ``heated'' part of the conversation) and Advocate B's speech disfluencies at $t$ are also more likely (the advocate could be mentally fatigued). Potential avenues forward include changing the unit of analysis to the entire conversational thread between the two target speakers or building extensions to the multiple mediator literature, i.e.~\citet{imai2013identification,vanderweele2014mediation,daniel2015causal,vanderweele2016mediation}. 

% One possible solution is to change the unit of analysis to be an entire conversational thread between the two (and only the two) target speakers. Yet, in multi-speaker conversations (such as the U.S. Supreme Court oral arguments) automatic thread disentanglement is a difficult and open problem in NLP \cite{uthus2013multiparticipant,elsner2010disentangling,mehri2017chat} and discarding threads that have more than two speakers changes the estimation sample. Alternatively, one could construct a causal model that incorporates information from previous utterances. However, the theory and empirical guarantees of estimating the effects of multiple causally dependent mediators is still very underdeveloped, and as \citet{daniel2015causal} note ``with more mediators, the complexity grows at such a rate that this [estimating path-specific effects] becomes impractical, even for three mediators.'' 

\textbf{Dependence between multiple language mediators.}
Our framework assumes one can computationally separate aspects of language.\footnote{This assumption is made in other NLP applications such as style transfer or machine translation \cite{prabhumoye2018style,li2018delete,hovy2020you}.} However, some sociolinguists argue aspects of language such as ``style'' cannot be separated from ``content'' because style originates in the content of people's lives and different ways of speaking signal socially meaningful differences in content \cite{eckert2008variation,blodgett2021sociolinguistically}.
If our mediator independence assumption (Eqn.~\ref{eqn:mediator-indep}) is violated, then we would have to turn to alternate estimation strategies to deal with this dependence.

\textbf{Dependence between social group perception and linguistic perception.}
Separating the direct and indirect causal paths in our framework relies on there being a \emph{decision-maker's latent  perception of social group} variable on the direct path between treatment and outcome and that this variable is independent from a \emph{decision-maker's latent perception of language} variable on the indirect path from treatment through mediators to outcome. However, ``indexical inversion'' considers ``how language ideologies associated with social categories produce the perception of linguistic signs'' \cite{inoue2006vicarious,rosa2017unsettling}.
Suppose Judge A perceives Advocate B as female, then Judge A might perceive Advocate B's language as more feminine even if it is linguistically identical to language used by male advocates. Furthermore, latent gender perception and latent language perception might interact in affecting the outcome through mechanisms such as rewarding ``conforming to gender norms''---an advocate who is perceived as a man might get penalized for using feminine language whereas an advocate perceived as a woman might get rewarded, e.g.~\citet{gleason2020beyond}.

\section{Conclusion} 
In this work, we specify a causal research design for \emph{differential treatment of social groups with language as a mediator}. We believe this research design is important for studying the direct and indirect causal effects in high-stakes decision making such as gender bias in the United States Supreme Court. Separating the indirect effect of treatment on outcome through interpretable language aspects allows us to estimate counterfactual queries about differential treatment when speakers use and do not use the same language.
Despite open theoretical and technical challenges, we remain optimistic that researchers can build upon this framework and continue to improve our understanding of decision makers' differential treatment of social groups.

% While empirically studying differential treatment is extremely important, the challenges and subtleties associated with this type of research design require careful thought on the part of the researcher. Causal identification with mediators relies on certain (often untestable) assumptions---such as sequential ignorability and conditional mediator independence---and estimation relies on fitting models for text, mediators, and outcome that have various levels of accuracy. Conceptualization of rich social constructs---such as gender or language---is difficult and they are often made of many constituent components. Once concepts are defined, there are often multiple operationalizations with tradeoffs between the specificity of the causal estimand and measurement difficulty. Additionally, the data for this research design will often consist of conversations and an open problem is dealing with the auto-dependence between temporal utterances. Yet, we remain optimistic that researchers can build upon this framework, address the open challenges, and continue to improve our understanding of differential treatment in high-stakes decision making settings.

\section*{Acknowledgments} 
The authors thank Abe Handler, Su Lin Blodgett, Sam Witty, David Jensen, and anonymous reviewers from the First Workshop on Causal Inference \& NLP for helpful comments. KK gratefully acknowledges funding from a Bloomberg Data Science Fellowship.  

\bibliography{bib}
\bibliographystyle{acl_natbib}

%Acknowledgements
%Su Lin, Abe, David Jensen 

% \appendix

% \section{Example Appendix}
% \label{sec:appendix}

% This is an appendix.

\end{document}